\documentclass[10pt, conference]{IEEEtran}

\usepackage{graphicx}
\usepackage{dblfloatfix}    

\usepackage[section]{placeins}

\usepackage[hyphens]{url} 
\usepackage{float}

\usepackage{amsmath}
\usepackage{amssymb}

\usepackage[ruled]{algorithm2e}
\usepackage[noend]{algpseudocode}
\usepackage{lipsum}
\usepackage{filecontents}
\usepackage{url}
\usepackage[para]{footmisc}
\usepackage{multirow}
\usepackage{caption}
\usepackage{subcaption}
\usepackage{tabularx}

\usepackage{pgfplots}
\usepackage{pgfplotstable}
\usepackage{siunitx}
\ifCLASSINFOpdf
\else
\fi

\graphicspath {{figures/}}

\pgfplotsset{grid style={dashed,gray}}
\pgfplotsset{minor grid style={dashed,red}}
\pgfplotsset{major grid style={dotted,green!50!black}}

\begin{document}
%


\title{A Foreground Inference Network for Video Surveillance Using Multi-View Receptive Field}

\author{\IEEEauthorblockN{Thangarajah Akilan}
\IEEEauthorblockA{Department of Electrical and Computer Engineering\\
University of Windsor, Windsor, Canada\\
Email: \{thangara\}@uwindsor.ca}}


\maketitle

\begin{abstract}
Foreground (FG) pixel labeling plays a vital role in video surveillance. Recent engineering solutions have attempted to exploit the efficacy of deep learning (DL) models initially targeted for image classification to deal with FG pixel labeling. One major drawback of such strategy is the lacking delineation of visual objects when training samples are limited. To grapple with this issue, we introduce a multi-view receptive field fully convolutional neural network (MV-FCN) that harness recent seminal ideas, such as, fully convolutional structure, inception modules, and residual networking. Therefrom, we implement a system in an encoder-decoder fashion that subsumes a core and two complementary feature flow paths. The model exploits inception modules at early and late stages with three different sizes of receptive fields to capture invariance at various scales. The features learned in the encoding phase are fused with appropriate feature maps in the decoding phase through residual connections for achieving enhanced spatial representation. These multi-view receptive fields and residual feature connections are expected to yield highly generalized features for an accurate pixel-wise FG region identification. It is, then, trained with database specific exemplary segmentations to predict desired FG objects.

The comparative experimental results on eleven benchmark datasets validate that the proposed model achieves very competitive performance with the prior- and state-of-the-art algorithms. We also report that how well a transfer learning approach can be useful to enhance the performance of our proposed MV-FCN.\\

\emph{keywords- Deep learning, Foreground-background clustering, Video-surveillance}
\end{abstract}

\IEEEpeerreviewmaketitle

\section{Introduction}
\begin{figure*}[!htb]
	\begin{center}
		\vspace*{0.0cm}
 		\includegraphics[width=140mm]{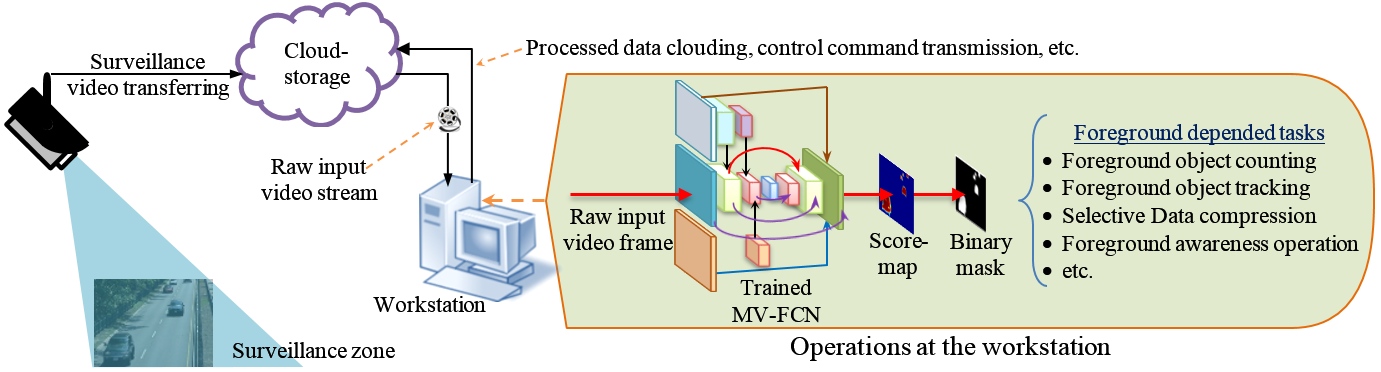}
	\end{center}
	\caption{A Probable Application Environment of the Proposed MV-FCN.}
	\label{fig:application_setup}
\end{figure*}
 
  Foreground region labeling is a crucial task in video surveillance used to detect moving objects in challenging conditions. It requires robust algorithms to tackle with varying environmental factors, like illumination changes and dynamic backgrounds \cite{Bouwmans2014}. It is also an integral part of various machine-vision problems, such as object segmentation \cite{Chen13}, \cite{aki2016}, \cite{AKILAN2018414}, image quality assessment \cite{Chow2016}, object discovery \cite{Kwak15}, visual tracking \cite{Zhou2016}, and human-robot/machine interaction \cite{guha16}. The primary objective of FG labeling is to place a tight mask on the most probable regions, in which moving objects mostly humans and vehicles can be identified. Such mask is, in many ways, very informative than a simple detection with bounding box as it allows close localization of objects, which is essential for retrieval, recognition, autonomous driving, and object preserved data compression for cloud-based systems \cite{Zhang2016InstanceLevelSF}. Besides, obtaining pixel-level foreground labels is also an important step towards general machine understanding of scenes. An example application setup is drawn in Fig.~\ref{fig:application_setup} to conceive the importance of this work.

%

Much attention has been paid to automate this process; and thus, there has been myriad of algorithms proposed that mainly include statistical approaches. For instance, Gaussian mixture models (GMM) \cite{Nguyen2015}, clustering algorithms, like conditional random field (CRF) \cite{ZouTIP2014} and graph-cut \cite{MaTIP2017}. However, some researchers focus on neural network (NN) models, like Self-Organizing Maps (SOM) \cite{ZhaoTIP2015} for this task. Here, a reasonable approach is to formulate it as a structured output problem that can be solved by training a system in an image-to-image fashion. This approach has been adopted in recent years' deep convolutional neural networks (DCNN/ deep convnets) for FG region labeling and gained wider acceptance.

One of the main challenges in DCNN-based methods is that dealing with objects of very different scales and the dithering effect at bordering pixels of FG objects. To deal with these challenges, we propose a new model inspired by Google introduced inception module \cite{SzegedyCVPR16} that performs convolution of multiple filters with different scales on the same input by simulating human cognitive processes in perceiving multi-scale information and Microsoft introduced ResNet \cite{He16} that acts as lost feature recovery mechanism. To enhance the knowledge of proposed network, we exploit intra-domain transfer learning that boosts the correct FG region prediction. Using this methodology is also inspired by human-like reasoning, in which the network learns new task precisely and more quickly by applying already learned knowledge, i.e., the weights and biases \cite{Wang:2011:AAAI}.

On a historical perspective, the theories of neural networks (NN) for the visual-based problems, arguably come from a pioneer computer vision system, the Mark I Perceptron machine by Rosenblatt in late 1950s \cite{BLOCK1970501}. Presumably, concurrent with that Hubel and Wiesel's ~\cite{Hubel68} discovery of neural connectivity pattern of cat's visual cortex, inspired Fukushima to introduce an NN referred to as Neocognitron~\cite{Fukushima80}, which is invariance to image translations. Later, the Neocognitron was devised with backpropagation mechanism that structures modern-day deep convnets. That is a multi-layered NN containing layers of several convolution, rectification, sub-sampling, and normalization operations. In which, the low-level convolutional layers operate as Gabor filters and color blob detectors~\cite{aki_smc17} that extract the information, such as, edges and/or textures, while the top-level layers provide the abstractive meaning of input data. The DCNNs have become the front-runner technology for various computer vision-based applications followed by Krizhevsky~\emph{et al.}'s \cite{Krizhevsky12} successful campaign with big achievements on the ImageNet 2012 Large-Scale Visual Recognition Challenge (ILSVRC-2012).

Subsequently, the deep CNNs have been effectively exploited for semantic segmentation/labeling \cite{ShelhamerPAMI17}, instance partitioning \cite{Zhang2016InstanceLevelSF}, and medical image sectioning \cite{Pereira16}, \cite{Ronneberger2015}. Thereupon we are interested in implementing a DCNN for the problem of FG object/region identification. Thus, the key insight of this paper is to propose a deep convnet that enhances feature learning for a better FG-region localization based on novel strategies introduced in the recent literature. We formulate this problem as a binary classification task with a DCNN, where the network is trained end-to-end with exemplary FG-BG segmentations to predict the most probable FG region for a given input frame. The proposed model is a multi-view receptive field fully convolutional neural network (MV-FCN) having two architectural phases: an encoder and a decoder that mainly combines inception modules and residual connections. Besides, the network is fully convolutional without any max pooling and fully connected layers. 

	\label{fig:overview}

In the following, we review the literature in Section~\ref{review} that covers sufficient CNN architectural information to understand the proposed model described in Section~\ref{Approach}. Sections \ref{experiments} presents details of the experimental setup and results along with discussion on compared FG detection algorithms. Finally, Section~\ref{conclusion} concludes the paper with future directions.

\section{Literature Review}\label{review}

\begin{figure*}[!htb]
	\begin{center}
		\vspace*{0.0cm}
		\includegraphics[width=180mm]{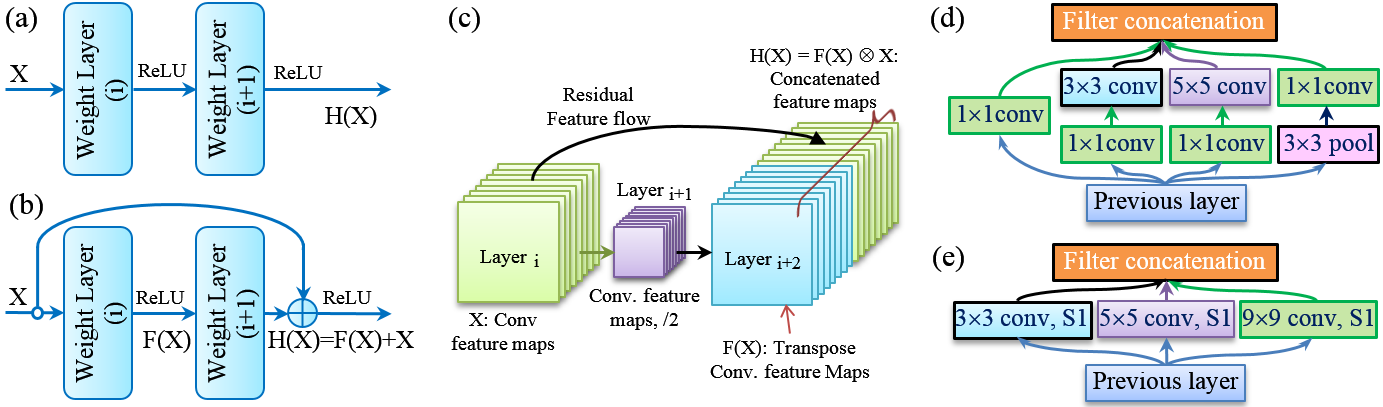}
	\end{center}
	\caption{ResNet and Inception Modules: (a) Standard CNN Connection, (b) ResNet Connection\cite{Ronneberger2015}, (c) Proposed Connection, (d) Original Inception Module\cite{DBLP:journals/corr/SzegedyLJSRAEVR14}  and (e) Our Inception Module.}
	\label{fig:feature_flows}
\end{figure*}

DCNNs have shown state-of-the-art performance over traditional methods, like GMM and graph-cut for the problem of FG detection/segmentation/localization. Here, the Fully convolutional networks \textbf{FCN} \cite{ShelhamerPAMI17} is a pioneer model that reinterprets the standard visual classification convnets as fully convolutional computation. It has been well exploited in many present-day applications, like sliding window-based detection \cite{ZhangICCV15}, \cite{Dai2016}, semantic segmentation \cite{icml2014c1_pinheiro14}, \cite{Caltagirone2017} and image retrieval/restoration \cite{EigenICCV13}, and spatial model for 3D face pose estimation \cite{TrigeorgisCVPR2017}. It is trained end-to-end and pixels-to-pixels \cite{ShelhamerPAMI17} using the whole image as input at a time. This model enhances pixel predictions at the last layer trough feature-level augmentation with a skip architecture that fuses the feature hierarchy to combine deep, coarse, semantic information and shallow, fine, appearance information from selected mid-layers. In contrast, our model does the coarse-level feature fusion in the inference path, like ResNet \cite{Ronneberger2015}.

In 2015, researchers from Microsoft introduced a CNN architecture with residual connections termed as \textbf{ResNet} that won the 1st place in the ILSVRC image classification competition with 3.57\% top-5 error. This network was built upon the philosophy of increasing depth of the network instead of widening, through residual connections to provide a better data representation. The ResNet architecture negates the vanishing gradient issue raises in deep networks by carrying important information in the previous layer to the next layer. Although such connection seems like an addition to the traditional CNN approach, it alleviates the training of the network and reduces number of parameters \cite{He16}. An illustration for the ResNet connection is given in Fig.~\ref{fig:feature_flows}~(b), where $X$ is input feature, $H(X)$ is any desired mapping, and $F(X)$ is a residual mapping. In \cite{He16}, the residual feature fusion operation $H(X) = F(X) + X$ is performed by a shortcut connection and element-wise addition. Contrastingly, our model stacks the futures depthwise as $H(X) = F(X) \bigotimes X$, like shown in Fig.~\ref{fig:feature_flows}~(c), where $\bigotimes$ denotes feature-map concatenation. This favors to have less number of filters in convolutional layers at the same time to carry forward earlier layer's features intact.

%
%

The \textbf{inception} module was a micro-architecture first introduced in \cite{DBLP:journals/corr/SzegedyLJSRAEVR14} by Szegedy~\emph{et al.}, following the success of ResNet \cite{SzegedyCVPR16}, \cite{CholletyCVPR17}. The module acts as computation of multiple filters with different scales on the same input as in Fig.~\ref{fig:feature_flows}~(d). Also, it performs average pooling at the same time. Finally, all the outcomes are aggregated along the channel dimension that to take advantage of multi-level feature representation, resulting in a higher discriminatory encoding. We simplify the inception connection with three different filters using stride of 1 ($S1$) convolution on the input image as shown in Fig.~\ref{fig:feature_flows}~(e). We also extract features through down-sampling operation on the three feature maps generated by the kernels and stack the features channel-wise at matching mid-layers through residual connections as depicted in Fig.~\ref{fig:network_detail}.


Researchers in \cite{Ronneberger2015} extend the FCN \cite{ShelhamerPAMI17} to function like an encoder-decoder CNN named \textbf{U-net} for an application of biomedical cell segmentation. In that, the activation maps after convolution at the encoding stage are concatenated with the activation maps at the decoding stage. Such structuring allows the network to exploit the original context information to supplement the features after up-sampling at the higher layers. In other words, it is a remedy for the lost spatial resolution due to max-pooling and striding at consecutive layers. The advantage of this model is that it works elegantly with less training samples and yields precise cell partitioning for 2D samples of Magnetic resonance imaging (MRI). Milletari~\emph{et al.}~\cite{Milletari16} extend this model to be trained end-to-end on MRI volumes to predict segmentation for the whole volume at once. They name the network V-net as it does volumetric medical image segmentation. The major difference of V-net from U-net is the volumetric convolutions as the input is a slice-wise volume (3D patches). Our model, on the other hand, has the following variations from the U-net: the traditionally deployed max-pooling operation in the contraction path achieves invariance feature but has a toll on localization accuracy \cite{ChenPAMI17}. Thus, we perform subsampling process by using 2D convolution with a stride rate of 2, a kernel size of $3\times3$, and zero padding. Hence, our model enhances feature learning through early and late stages micro inception modules.

\section{Proposed MV-FCN Architecture}\label{Approach}
\begin{figure*}[!htb]
	\begin{center}
		\vspace*{0.2cm}
		\includegraphics[width=180mm]{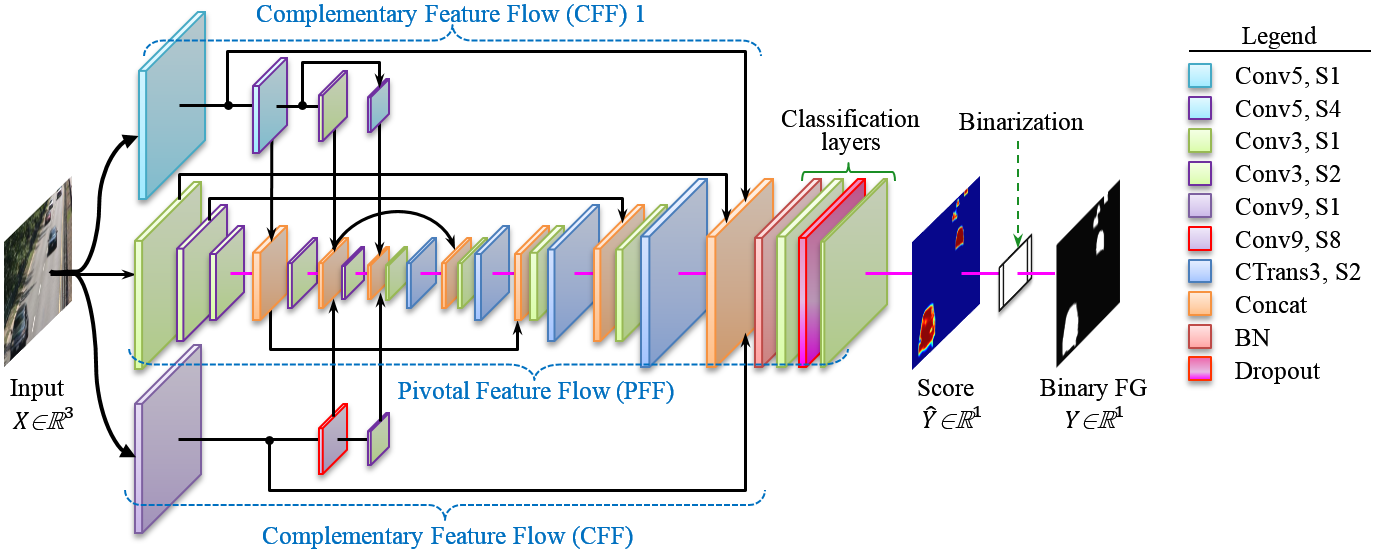}
	\end{center}
	\caption{Schematic Diagram of the MV-FCN: Conv$k$, S$i$, CTrans$k$, Concat, and BN stand for convolution using kernel size of $k$ and stride of $i$, transpose convolution with filter size of $k$, activation maps concatenation, and batch normalization operations, respectively.}
	\label{fig:network_detail}
\end{figure*}

Figure~\ref{fig:network_detail} abstracts away details of the proposed multi-view receptive field fully convolutional network (MV-FCN) through a schematic. The MV-FCN integrates two complementary feature flows (CFF) and a pivotal feature flow (PFF). The PFF is essentially an encoder-decoder CNN while CFF1 and CFF2 complement its learning ability. The PFF only uses convolution kernels size of $3\times3$, while CFF1 and CFF2 utilize filters size of $5\times5$ and $9\times9$ respectively in their first conv layers. However, after their first sub-sampling convolutional operations they use filter size of $3\times3$ in their subsequent layers, so their output activation maps match a middle layer in the PFF for a successful feature-level augmentation. Thus, the features learned in the complementary layers are merged with appropriate intermediate feature maps in the PFF through residual connections Using such heterogeneous convolutional kernels \textcolor{blue}{captures information available from different scales and} provides both local and global context \cite{badrinarayanan2017segnet} and the fusion of feature maps from encoding layers that hold high-frequency detail resulting to sharper foreground boundaries.

In the encoding phase of PFF, four convolutional layers are networked sequentially after the very first conv layer that generates 16 channels with spatial dimension same as the input. Each of the four conv layers performs spatial subsampling by using a kernel size $3\times3$ with stride of 2 such that the encoding process outputs activation map with a dimension of $15\times20\times96$. The decoding phase, on the other hand, employs four transpose convolutional layers interspace with residual feature concatenations and regular conv layers. Consequently, the decoding stage ends up with an inception module (layer 28 in Table~\ref{table:network_detail}) that merges the first stage activations from PFF, CFF1, and CFF2 with the final stage decoding activations resulting to a feature map of $240\times320\times112$.

The extracted features from various conv layers in the encoding path are also combined with spatially matching up-sampled feature maps in the decoding path systematically. As stated earlier in section~\ref{review}, this strategy is an elegant solution for the lost of spatial resolution due to series of subsampling and convolutional operations carried out over the encoding process \cite{Ronneberger2015}. Hence, all the convolutions are immediately followed by ReLU activation functions, except the transpose convolution (it is generally referred as deconvolution) and the final layer. Top classification layers consist of a batch-normalization, conv with 128 channels followed by drop out of 0.3, and finally a single channel output conv with Sigmoid activation function. Table~\ref{table:network_detail} summarizes the network detail, where conv2D and conv2DT denote 2D convolution and its transpose, respectively. The integers in the parentheses in layer type refer the kernel size and stride rate in the order while the $None$ in output shape refers the mini-batch size. In total, the proposed model takes 494,337 trainable parameters.

\begin{table}[!pt]
	\begin{minipage}{\columnwidth}
	\begin{center}
	\vspace*{0.0cm}
	\begin{tabular}{@{}|c|c|c|c|@{}}
	\hline
	\multirow{2}{*}{Layer ID} & \multirow{2}{*}{Layer type} & \multirow{2}{*}{Output Shape} & \multirow{2}{*}{\shortstack{Input \\(layer ID)}}\\
	& & & \\
	\hline\hline
	1 & Input Layer & (None, 240, 320, 3) & mini-batch\\
	2 & Conv2D (3, 1) & (None, 240, 320, 16) & 1\\
	3 & Conv2D (5, 1) & (None, 240, 320, 16) & 1\\
	4 & Conv2D (9, 1) & (None, 240, 320, 16) & 1\\
	5 & Conv2D (3, 2) & (None, 120, 160, 16) & 2\\
	6 & Conv2D (3, 2) & (None, 60, 80, 32) & 5\\
	7 & Conv2D (5, 4) & (None, 60, 80, 32) & 3\\
	8 & Concatenation & (None, 60, 80, 64) & 6, 7\\
	9 & Conv2D (3, 2) & (None, 30, 40, 32) & 8\\
	10 & Conv2D (3, 2) & (None, 30, 40, 32) & 7\\
	11 & Conv2D (9, 8) & (None, 30, 40, 32) & 4\\
	12 & Concatenation & (None, 30, 40, 96) & 9, 10, 11\\
	13 & Conv2D (3, 2) & (None, 15, 20, 32) & 12\\
	14 & Conv2D (5, 4) & (None, 15, 20, 32) & 7\\
	15 & Conv2D (3, 2) & (None, 15, 20, 32) & 11\\
	16 & Concatenation & (None, 15, 20, 96) & 13, 14, 15\\
	17 & Conv2D (3, 1) & (None, 15, 20, 64) & 16\\
	18 & Conv2DT (3, 2) & (None, 30, 40, 64) & 17\\
	19 & Concatenation & (None, 30, 40, 160) & 18, 12\\
	20 & Conv2D (3, 1) & (None, 30, 40, 32) & 19\\
	21 & Conv2DT (3, 2) & (None, 60, 80, 32) & 20\\
	22 & Concatenation & (None, 60, 80, 96) & 21, 8\\
	23 & Conv2D (3, 1) & (None, 60, 80, 32) & 22\\
	24 & Conv2DT (3, 2) & (None, 120, 160, 16) & 23\\
	25 & Concatenation & (None, 120, 160, 32) & 24, 5\\
	26 & Conv2D (3, 1) & (None, 120, 160, 32) & 25\\
	27 & Conv2DT (3, 2) & (None, 240, 320, 64) & 26\\
	28 & Concatenation & (None, 240, 320, 112) & 27, 2, 3, 4\\
	29 & BatchNorm & (None, 240, 320, 112) & 28\\
	30 & Conv2D (3, 1) & (None, 240, 320, 128) & 29\\
	31 & Dropout & (None, 240, 320, 128) & 30\\
	32 & Conv2D (3, 1) & (None, 240, 320, 1) & 25 \\

\hline \hline
\end{tabular}
\end{center}
\caption{Layer detail of the MV-FCN.}
\label{table:network_detail}
\end{minipage}
\end{table}

In summary, the MV-FCN does not employ max pooling or hidden fully connected (FC) layers, but subsumes convolutional (conv), transpose convolutional (CTrans), and symmetric expanding paths with inception and residual connections to capture contextual information for an accurate FG inferencing. The network is capable of taking any spatial dimensions of input images and resize them into $240\times320$ by using nearest-neighbor scaling algorithm to match with the input layer dimension. The convolutional layers use stride rate of 1 in all directions, except the sub-sampling layers that perform convolution with a stride rate of $k-1$, where $K$ is the kernel size. At this juncture, it is vital to discuss about the intricacies of the core components and the functions utilized in the proposed network.

\subsection{Convolutional Layer}
The convolutional layer is the core unit of modern deep learning architectures that is determined by its kernel weights that are updated during training via backpropagation. All the filter weights are fixed like a system memory; some literature refer it as anchor vectors since they serve as reference visual patterns in the testing phase. Output feature map $C$ w.r.t. a kernel $\mathbf{\omega}$, its associated bias $b$, and an input image/patch $\mathbf{x}$ the convolutional operation is performed as
\begin{equation} \label{conv}
C(m,n) = b + \mathop \sum \limits_{k = 0}^{K - 1} \mathop \sum \limits_{l = 0}^{K - 1} \omega(k,l)\ast \mathbf{x}(m + k,n + l),
\end{equation}
where $\ast$, $K$, $\{m,n\}$, and $\{k, l\}$ represent the convolutional operation, size of the kernel, first coordinate or origin of the image/patch, and element index of the kernel respectively. Hence, feature map dimension of the conv layer is given by $(I_s - K_s + 2\times P)/S + 1$, where $I_s, K_s, P$, and $S$ denotes size of input image/path, filter size, number of zero-padded pixels, and stride rate respectively.

\subsection{Transpose Convolution}
\begin{figure}[!hb]
	\begin{center}
		\vspace*{0.0cm}
		\includegraphics[width=80mm]{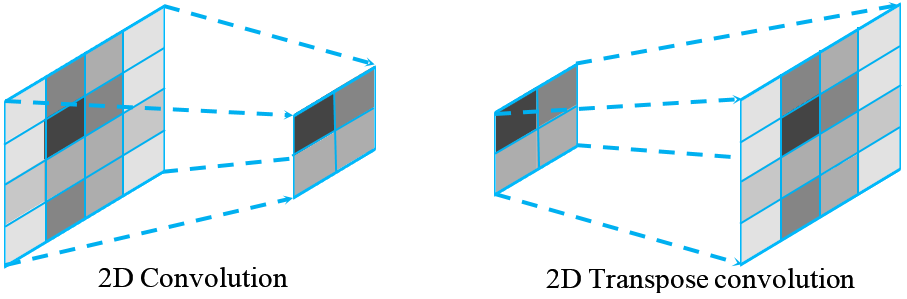}
	\end{center}
	\caption{Illustration of convolution and its transpose operations: A 2D conv with $K=3, S=2,$ and $P=1$, and its corresponding transpose conv with $K'=K, S'=1,$ and $P'=K-P-1$~\cite{odena2016deconvolution}.}
	\label{fig:conv_deconv}
\end{figure}

Transpose convolutional layers perform up-sampling, i.e., the transpose/gradient of 2D convolution such that its output spatial dimension becomes twice as the input and dense activation map, as illustrated in Fig.~\ref{fig:conv_deconv} without losing the connectivity pattern. Although it looks like an image resizing process, it has trainable parameters for the up-sampling stage, and these parametric quantities are updated during training. It is achieved by inserting zeros between consecutive neurons in the input receptive field, then sliding the convolutional kernel with unit strides \cite{dumoulin2016guide}. To elaborate it, if a desired convolution is governed by kernel size of $K$, stride rate of $S$, zero padding size of $P$, and its output has size of $i'$ then the associated transpose convolution can be computed with such a kernel $K'=K$, stride $S'=1$, padding $P'=K-P-1$, $\tilde{i'}$, and $\alpha$, where $\tilde{i'}$ is the size of the diluted input obtained by imputing $S-1$ zeros between each input neuron, and $\alpha=((i+2P-K)\hspace{0.25cm} mod\hspace{0.25cm} S)$ represents the number of zeros inserted to the top and right edges of the input that results an output feature size of:
\begin{equation}
O' = S(i'-1) + \alpha + K - 2P.
\end{equation}

The modern deep neural networks implemented for image-to-image learning use multiple layers of transpose conv for generating images or feature maps from a series of lower resolution descriptions \cite{odena2016deconvolution}. This idea has a long history in signal processing domain, originally developed for the efficient computation of the undecimated wavelet transform (UWT) known as `algorithme a trous' \cite{Holschneider1990}.
More detail and computational information on transpose convolutional operation can be referred from \cite{odena2016deconvolution} and \cite{dumoulin2016guide}.

\subsection{Activation Functions}
The adaptation of activation functions in neural networks (NN) can be referenced to the work of McCulloch and Pitts in late 1943 \cite{McCulloch1943}, where the activation function rectify the input to either $1$ or $-1$ if its value is positive or negative, respectively.

\textbf{ReLU}s are nonlinear activations generally used after convolutional operations. It can be formally defined as in (\ref{eqn:relu}) when taken a case where there are $K$ number of anchor vectors, denoted by $\mathbf{w}_k \in \mathbb{R}^N, k=1,2,\dots, K$. For a given input $\mathbf{x}$, the correlations with $\mathbf{a}_k$ and $k=1,2,\dots, K$, defines a nonlinear rectification to an output $\mathbf{y} = (y_1, \dots, y_K)^T$, where

\begin{equation} \label{eqn:relu}
y_k(\mathbf{x, a}_k) = \max(0, \mathbf{a}_k^T\mathbf{x}) \equiv \emph{ReLU}(\mathbf{a}_k^T\mathbf{x}),
\end{equation}
i.e., it clips negative values to zero while keeping positive ones intact. The benefit of ReLU is sparsity, overcoming vanishing gradient issue, and efficient computation than other activations.

\textbf{Sigmoid} function, on the other hand, has output in the range $[0, 1]$ for an input $\mathbf{x}$ and it is defined by

\begin{equation} \label{eqn:sigmoid}
 f(\mathbf{x}) = \frac{1}{1 + exp(-\mathbf{x})}.
\end{equation}
Therefore, it is very appropriate for binary classification tasks, like in this work and linear regression problems. A thorough exposition of the purpose of activation functions in NN with graphical examples can be found in \cite{Kuo2016406}.

\subsection{Batch normalization}
The batch normalization (layer 29 in Table~\ref{table:network_detail}) before the final classification layer has multifaceted benefits: (i) reducing internal covariate shift - During training, there is a change in the distribution of activation maps as network parameters are being tuned. Such condition challenges the learning, but the BN alleviate pressure by maintaining the mean and standard deviation of the activation close to 0 and 1, respectively. (ii) Effect of regularization - Since the batch of examples given in the training are normalized, it increases the generalization of the model. It is also claimed that BN allows to reduce the strength of dropout. (iii) Counterbalancing vanishing or exploding gradients - When the BN is located prior to non-linearity, it avoids an undesirable situation, where the training saturates areas of non-linearities, solving the issues of vanishing exploding gradients.

Mathematically it can be defined as follows. Let the output of a layer $\mathbf{X}\in\mathbb{R}^{N,D}$, where $N$ is the number of samples available in the mini-batch and $D$ is the number of hidden neurons, then normalized matrix $\hat{\mathbf{X}}$ is given as in (\ref{eqn:BN}) \cite{ioffe2015batch}.

\begin{equation}\label{eqn:BN}
	\hat{\mathbf{X}} = \frac{\mathbf{X} - \mu_B}{\sqrt{\sigma_B^2 + \epsilon}},
\end{equation}
where $\mu_B, \sigma_B^2$, and $\epsilon$ refer to the mean and variance of the mini-batch, and a small value of 0.001 to prevent division by zero, respectively. Then the layer maintains its representational strength by testing the identity transform as:
\begin{equation}\label{eqn:identity_transform}
y = \gamma \hat{X} + \beta,
\end{equation}
where, $\beta$ and $\gamma$ are trainable parameters that are initialized with $\beta = 0$ and $\gamma = 1$, in this work. Note that, when $\beta = \mu_B$ and $\gamma = \sqrt{\sigma_B^2 + \epsilon}$ Eqn.~(\ref{eqn:identity_transform}) returns the previous layer's activation map.

\subsection{Training strategy}

\begin{table}[!ht]
	\begin{minipage}{\columnwidth}
		\begin{center}
			\vspace*{0.0cm}
			\resizebox{1.0\columnwidth}{!}{%
				\begin{tabular}{@{}|c|c|c|c|@{}}
					\hline
					\multirow{2}{*}{Dataset} &  \multirow{2}{*}{\shortstack{Frame size \\($W\times H$)}} & \multirow{2}{*}{Nature} & \multirow{2}{*}{$N$ frames}\\
					& & &\\
					\hline\hline
					Highway & $320\times240$ & \multirow{2}{*}{Baseline} & 1229 \\
					Office & $360\times240$  &                           & 1447 \\
					\hline
					Canoe & $320\times240$ & \multirow{3}{*}{\shortstack{Dynamic \\background}} & 342 \\
					Boats & $320\times240$  &                                    & 6026 \\
					Overpass &  $320\times240$ &                             & 440 \\
					\hline
					Traffic &  $320\times240$ &  \multirow{2}{*}{\shortstack{Camera jitter}} & 609 \\
					Boulevard &  $320\times240$ &                                 & 1004 \\
					\hline
					CopyMachine &  $720\times480$ & \multirow{2}{*}{\shortstack{Shadow}} & 1401 \\
					PeopleInShade &  $380\times244$ &  & 829 \\
					\hline
					TwoPositionPTZCam &  $570\times340$ & \multirow{1}{*}{\shortstack{PTZ camera}} & 449 \\
					\hline
					Turnpike\_0\_5fps &  $320\times240$ & \multirow{1}{*}{\shortstack{Low Framerate}} & 350 \\
					\hline \hline
				\end{tabular}}
		\end{center}
		\caption{Dataset Summary.}
		\label{table:dataset_summary}
	\end{minipage}
\end{table}

\textbf{Exclusive sets:} 
\begin{figure}[!ht]
	\begin{center}
		\vspace*{0.0cm}
		\includegraphics[width=85mm]{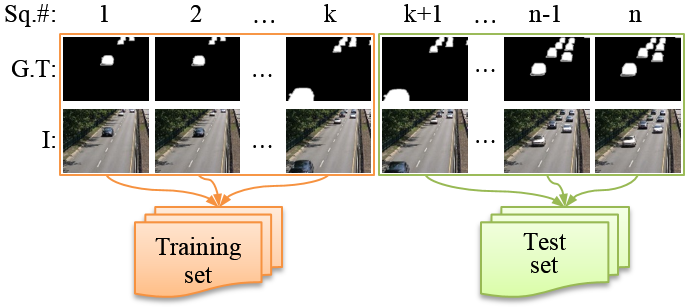}
	\end{center}
	\caption{Ordered Exclusive Split of Training and Test Sets: G.T- ground truth, I- RGB raw input image, S.q\#- sequence ID.}
	\label{fig:data_split}
\end{figure} 
We target the widely used benchmark database the change detection 2014~\cite{Wang2014}. Table~\ref{table:dataset_summary} briefs the properties of the datasets. Since the database has less number of annotated ground truths, where both the FG and BG are presented in the same frame, we employ data augmentation by applying random transformations with rotation within 10 degrees, translation vertically and horizontally with a fraction of 0.1 from the total height and width, and zooming in range of 0.1 inside image samples. These data augmentations are done on training images and the corresponding ground truths during training. Naturally, this allows the network to learn invariance to such transformations, without a need to see these variant samples in the annotated benchmark datasets. To form exclusive sets of training and test data, the available samples are divided in sequence order, whereby training set takes 70\% while test set takes 30\% of the total number of samples that have ground truths with FG and BG information in a particular dataset. This way of data splitting is more appropriate rather than a random selection since the images in the datasets are frames from video sequences. Because, a random choice of samples may pick a $frame_{t}$ for training set while picking a temporally closest frame, like $frame_{t+1}$ or $frame_{t-1}$ for test set. There can be many such instances in random selection resulting in mere exclusiveness of training and test sets. Figure~\ref{fig:data_split} demonstrates the data split used in this work, in which $n$ is the total number of samples that has ground truths in the sequence and $k = \lfloor n\times0.7\rfloor$ that is the dividing point (frame no.) for the ordered split.   \\

\textbf{Optimizer:} The MV-FCN is trained by using Adam-optimizer that minimizes binary cross-entropy loss defined by (\ref{eqn:cost_fun}), where optimizer takes a base learning rate of $0.0002$ \textcolor{blue}{with a learning rate scheduler that reduces the learning rate by factor of 0.8 over the training.}


\begin{equation} \label{eqn:cost_fun}
\resizebox{0.9\columnwidth}{!}{$E = \frac{-1}{n} \sum\limits_{n=1}^N \left[ p_n \log \hat{p}_n + (1 - p_n) \log(1 - \hat{p}_n) \right]$},
\end{equation}
where it takes two inputs; first one is the output from the final layer of the network (layer 32 in Table~\ref{table:network_detail}) with dimension of $N \times C \times H \times W$, which maps the FG pixel probabilities $\hat{p}_n = \sigma(x_n) \in [0, 1]$ using Sigmoid non-linearity function $ \sigma(.)$ defined earlier in Eqn.~\ref{eqn:sigmoid}. And the second one is target $p_n \in [0, 1] $ with the same dimension as the first one, where $N, C, H$, and $W$ represent the batch size, the number of channels, hight, and width respectively of the inputs. \textcolor{blue}{In this case, $p_n$ is the ground truth segmentation images whose pixel values are normalized.} The network is trained on each video sequence separately.     \\

\textbf{Transfer learning:} 
\begin{table}[!ht]
	\begin{minipage}{\columnwidth}
		\begin{center}
			\vspace*{0.0cm}
				\begin{tabular}{@{}|c|c|@{}}
					\hline
					{Model Fine-tuned for} & {Model Transferred from}\\
					\hline\hline
					Highway & Turnpike\_0\_5fps \\
					Office & CopyMachine \\

					Canoe & Boats \\
					Boats & Canoe \\
					Overpass &  Pedestrians \\
					
					Traffic &  Highway \\
					Boulevard &  TwoPositionPTZCam \\
				
					CopyMachine &  Office \\
					PeopleInShade &  Pedestrians \\
				
					TwoPositionPTZCam &  Turnpike\_0\_5fps \\
					
					Turnpike\_0\_5fps &  TwoPositionPTZCam \\
					\hline \hline
			\end{tabular}
		\end{center}
		\caption{Transfer Learning Dataset Pairs.}
		\label{table:trasfer_learning_detail}
	\end{minipage}
\end{table}
To improve the network's learning experience we incorporate intraclass domain transfer. Table~\ref{table:trasfer_learning_detail} lists the fine-tuning dataset pairs. For instance, the pre-trained network with \emph{TwoPositionPTZCam} is fine-tuned for \emph{Turnpike\_0\_5fps}. Here, both the domain have moving vehicles as FG objects. The theoretical and philosophical expositions of transfer learning can be found in \cite{Weiss2016} and  \cite{Pan_2010_transferlearning}.  \\

\textbf{Training Environment:} Python with Keras libraries (Tensorflow backend) is used as a software platform for the implementation of the model. The network is then mainly trained on a GeForce GTX 1060-6 GB GPU with Intel(R) Core(TM) i7-4770 CPU @ 3.40 GHz and 32 GB memory (RAM). In average the training takes about 2 hours on the GPU for each dataset when batch size is 8 and maximum of 30 epochs. The testing is purely carried out on CPU and it takes about 0.445 second per sample in average.
                
\subsection{Binary Foreground Mask}
\begin{figure}[!ht]
	\begin{center}
		\vspace*{0.0cm}
		\includegraphics[width=89mm]{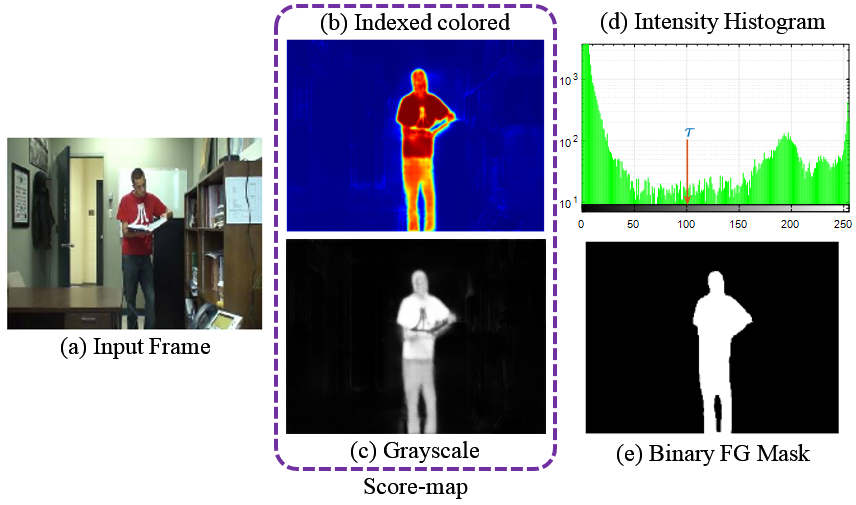}
	\end{center}
	\caption{Creating FG Mask: Applying an appropriate threshold to the score-map generated from the last classification layer of MV-FCN for a frame taken from the Office dataset.}
	\label{fig:score_to_FG}
\end{figure}

It is also crucial to create a binary mask that localizes the interested FG region in a given frame. We apply a threshold to the score-map generated by the trained MV-FCN at frame-level to form a FG mask like shown in Fig.~\ref{fig:score_to_FG}, where the threshold $\tau$ is a dataset-specific global parameter set empirically in the range $[0.05, 0.75]$. Then to clean noisy artifacts, we post process the thresholded binary image through neighborhood pixel connectivity that removes regions with less than 50 pixels. In another approach, We employ the Otsu's clustering-based model to choose appropriate threshold automatically, since the score-map is a representation of bi-modal image. Otsu's algorithm iteratively finds a threshold $\tau$ that lies in between two peaks of the intensity histogram such that the intra-class variances of FG and BG classes are minimum. There, the weighted sum of within-class variances is defined as:
\begin{equation}\label{eqn:otsu}
\sigma_\rho^2(\tau) = \rho_0(\tau)\sigma_0^2(\tau) + \rho_1(\tau)\sigma_1^2(\tau),
\end{equation}
where the weights $\rho_0$ and $\rho_1$ are the probabilities of BG and FG classes clustered by a threshold $\tau$, and the variances of these two classes are $\sigma_0^2$ and $\sigma_1^2$ respectively. An explicit derivation of the method can be found in \cite{otsu1979threshold}. Note that the binarization process is not part of the MV-FCN training procedure, but exclusive for testing stage as the numerical analysis is made on the binary masks.

\section{Experimental Setup, Results, and Discussion}\label{experiments}
To provide a better understanding of the performance, we select eleven various sequences from the change detection database \cite{Wang2014} that consists of diversified change and motion, including benchmarks of baseline, dynamic background, camera jitter, shadow, videos shot with PTZ camera, and low framerate. A succinct description of the datasets is given in Table~\ref{table:dataset_summary}. Hence, the general nature of the datasets as follows. The \textbf{baseline} benchmark represents a mixture of mild challenges, like subtle background motion, isolated shadows, swaying tree branches, and natural illumination changes. The \textbf{dynamic background} category includes scenes with strong (parasitic) BG motion: boats and canoes on shimmering water, or a man walking on a shore of a shimmering water body. The \textbf{camera jitter} datasets contain outdoor videos captured by vibrating cameras due to high wind and unstable mount. The jitter magnitude varies from one video to another. The \textbf{shadow} category comprises indoor video exhibiting strong as well as faint shadows. Here, some shadows are cast by moving objects. Lastly, in \textbf{PTZ camera recordings}, adjustments in camera strongly changes the backgrounds of a recorded video. Such conditions break the assumption of traditional BG modeling algorithms that the recording devices are relatively static, or move slowly, and thus it challenges the most algorithms.

\subsection{Qualitative Analysis}

\begin{figure*}[!ht]
	\begin{center}
		\includegraphics[width=175mm]{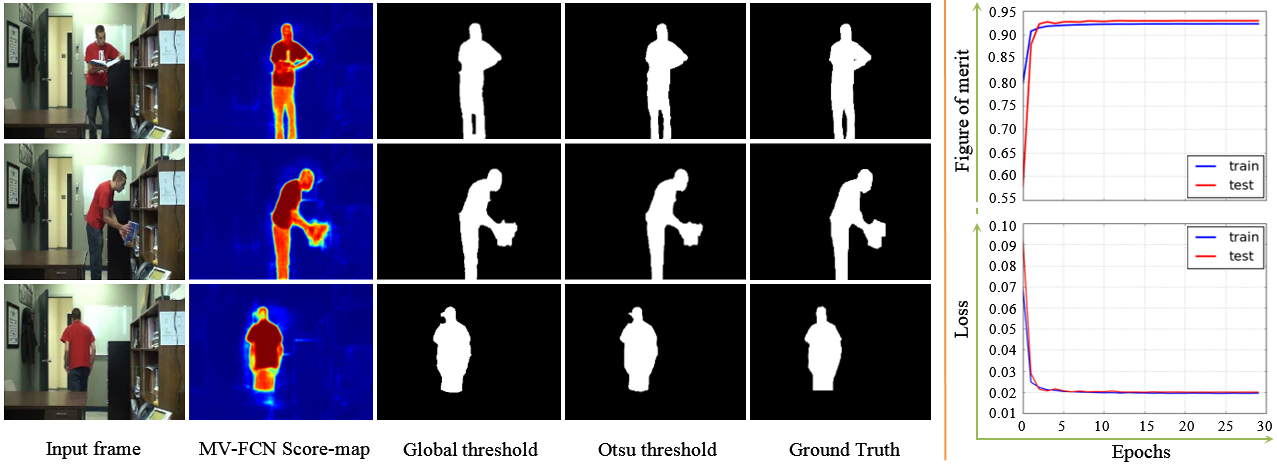}
	\end{center}
	\caption{Sample results for the Office dataset. Col. 1-5: Sample input frames, MV-FCN generated score-maps, binary FG masks with empirical and Otsu's thresholds. Col. 6: training and validation FoM and loss respectively in the top and bottom.}
	\label{fig:office_dataset_samples}
\end{figure*}
\begin{figure*}[!ht]
	\begin{center}
		\includegraphics[width=175mm]{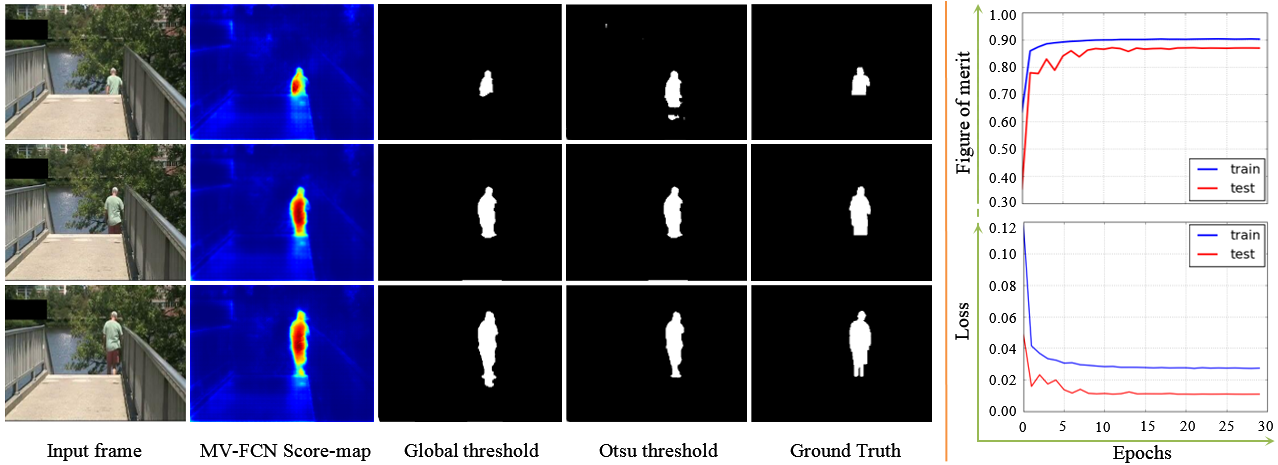}
	\end{center}
	\caption{Sample Results for the Overpass Dataset. Col. 1-5: Sample input frames, MV-FCN generated score-maps, binary FG masks with empirical and Otsu's thresholds. Col. 6: Training and validation FoM and loss respectively in the top and bottom.}
	\label{fig:overpass_dataset_samples}
\end{figure*}



\begin{figure*}[!htb]
	\begin{center}
		\includegraphics[width=175mm]{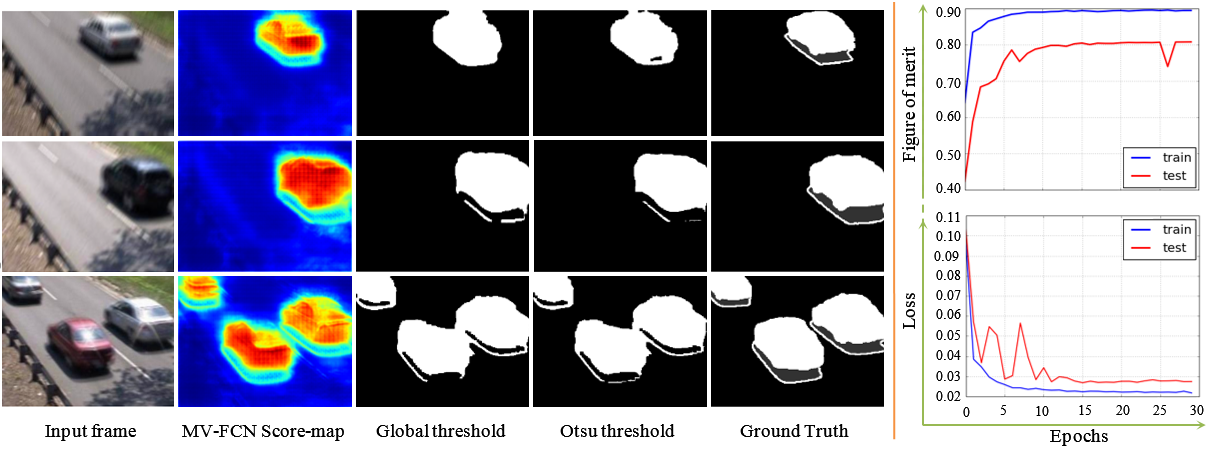}
	\end{center}
	\caption{Sample Results for the Traffic Dataset. Col. 1-5: Sample input frames, MV-FCN generated score-maps, binary FG masks with empirical and Otsu's thresholds. Col. 6: Training and validation FoM and loss respectively in the top and bottom.}
	\label{fig:traffic_dataset_samples}
\end{figure*}

\begin{figure*}[!htb]
	\begin{center}
		\includegraphics[width=175mm]{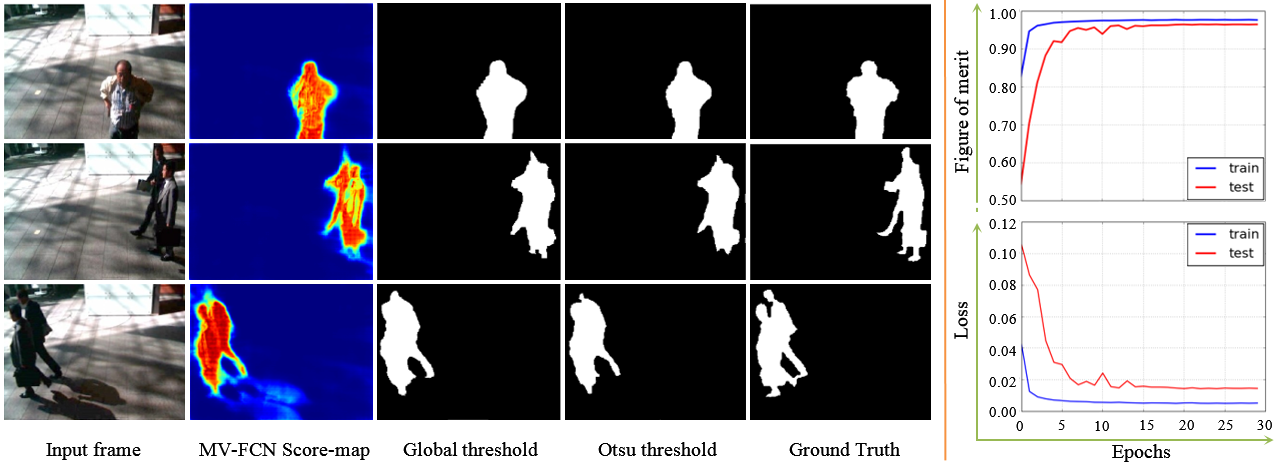}
	\end{center}
	\caption{Sample Results for the PeopleInShade Dataset. Col. 1-5: Sample input frames, MV-FCN generated score-maps, binary FG masks with empirical and Otsu's thresholds. Col. 6: Training and validation FoM and loss respectively in the top and bottom.}
	\label{fig:PeopleInShade_dataset_samples}
\end{figure*}

\begin{figure*}[!htb]
	\begin{center}
		\includegraphics[width=175mm]{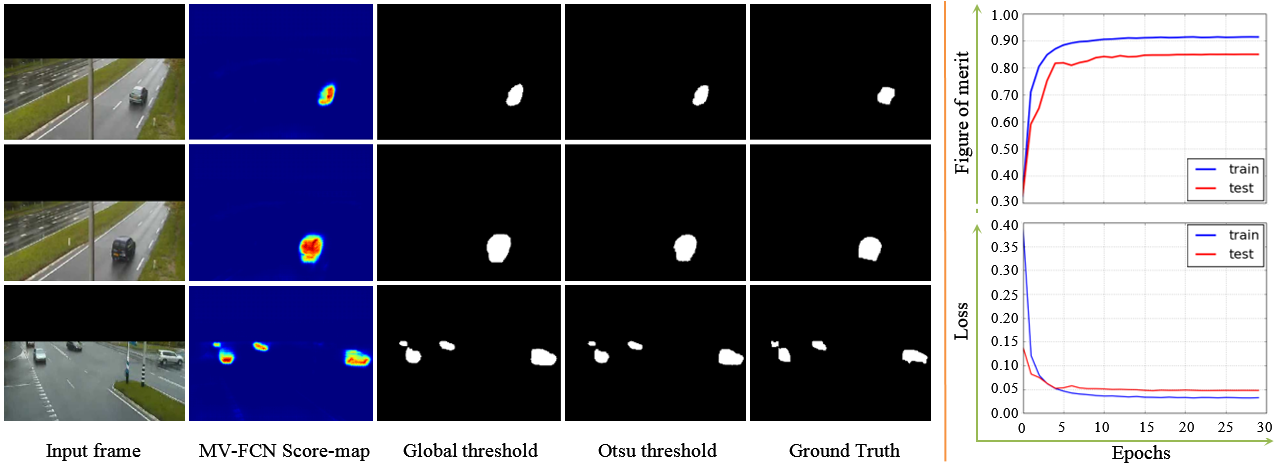}
	\end{center}
	\caption{Sample Results for the TwoPositionPTZCam Dataset. Col. 1-5: Sample input frames, MV-FCN generated score-maps, binary FG masks with empirical and Otsu's thresholds. Col. 6: Training and validation FoM and loss respectively in the top and bottom.}
	\label{fig:twoPositionPTZCam_dataset_samples}
\end{figure*}

\begin{figure*}[!ht]
	\begin{center}
		\includegraphics[width=175mm]{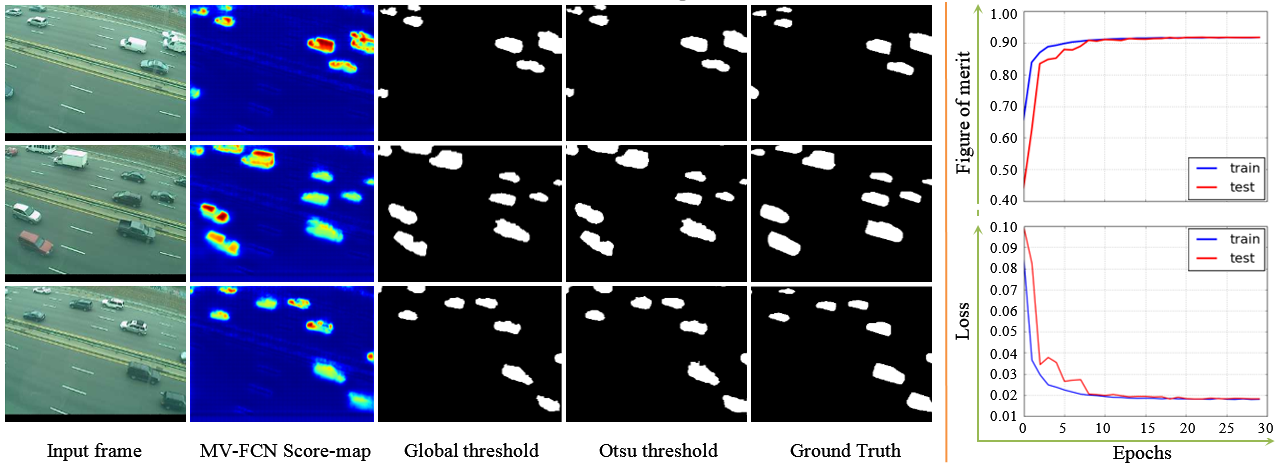}
	\end{center}
	\caption{Sample Results for the Turnpike\_0\_5fps Dataset. Col. 1-5: Sample input frames, MV-FCN generated score-maps, binary FG masks with empirical and Otsu's thresholds. Col. 6: Training and validation FoM and loss respectively in the top and bottom.}
	\label{fig:turnpike_0_5fps_dataset_samples}
\end{figure*}

\begin{figure}[!ht]
	\begin{center}
		\includegraphics[width=88mm]{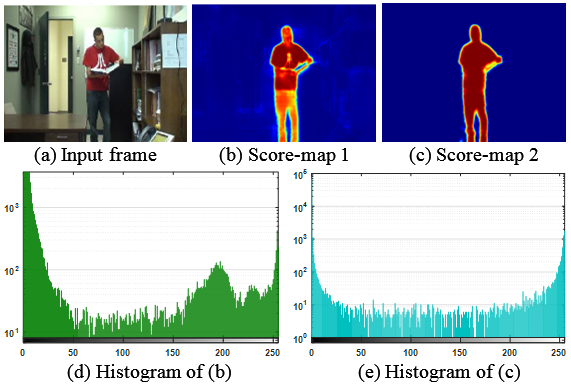}
	\end{center}
	\caption{MV-FCN score map when: (b) trained from scratch, (c) fine-tuned with intra-class transfer learning.}	 \label{fig:TL_imapat}
\end{figure}

To limit the number of pages of this report, three sample results\footnote{Rest of the results will be available in the project web page.} from a selected video sequence per category from Table~\ref{table:dataset_summary} are shown in
Fig.~\ref{fig:office_dataset_samples} - Fig.~\ref{fig:turnpike_0_5fps_dataset_samples}. Hence, The impact of intraclass transfer learning is visualized using one sample from the Office dataset in Fig.~\ref{fig:TL_imapat}. Such transfer learning technique allows the network to produce stronger discrimination of FG regions from BG as the distribution of probability falls around two distinct peaks, generally with the intensity values of $0$ (dark as BG) and $255$ (bright as FG). Visual results of proposed MV-FCN appear close to the ground-truth references; however, it has to be quantitatively analyzed for further validation. The following subsection~\ref{numerical_analysis} provides the numerical analysis in terms of Figure-of-Merit (FoM).

\subsection{Quantitative Analysis: FoM} \label{numerical_analysis}
The goal of FG inferencing is to label the pixels of a given video frame as being part of an expected FG object or not (i.e., BG). In such problem domain, the standard performance measure used is Figure-of-merit or FoM in short. The FoM measures the similarity between the predicted FG region and the ground-truth for a concerned FG object present in the image, and is defined as a weighted harmonic mean measure of recall and precision, i.e., a region of intersection divided by the union of predicted and actual FG regions. It is also referred as intersection-over-union (IoU) as in (\ref{eqn:fom}).

\begin{equation}\label{eqn:fom}
    FoM = \frac{2\times(Precision\times Recall)}{Precision + Recall}
    \footnote{\resizebox{\columnwidth}{!}{${\frac{2\times TP/(TP+FN)\times TP/(TP+FP)}{TP/(TP+FN) + TP/(TP+FP)} = \frac{2\times TP^2}{TP(TP+FP+TP+TN)}=\frac{2\times TP}{(TP+FP)+(TP+FN)}}$}},
\end{equation}
where recall is the detection rate defined by $TP/(TP + FN)$ and precision is the percentage of correct prediction compared to the total number of detections as positives, given by $TP/(TP + FP)$, where $TP, FN$, and $FP$ refer true positive, false negative, and false positive respectively. For a given output $X$ from the proposed MV-FCN, i.e., the probabilities over a set of pixels $V = \{1, 2,\cdots, N\}$ in the input image, and $Y \in \{0, 1\}$ the ground-truth assignment for the set $V$, where 0 and 1 refer the BG and FG object pixels respectively, then (\ref{eqn:fom}) can be formalized as (\ref{eqn:iou}).

\begin{equation}\label{eqn:iou}
    FoM = \frac{2\times I(X)}{U(X)},
\end{equation}
where $I(X)$ and $U(X)$ can be approximated as follows:

\begin{equation}\label{eqn:iofx}
    I(X) = \sum_{v\in V}X_v * Y_v + \epsilon \equiv TP,
\end{equation}

\begin{equation}\label{eqn:uofx}
    U(X) = \sum_{v\in V}(X_v + Y_v) + \epsilon \equiv (TP+FP) + (TP+FN),
\end{equation}
where $\epsilon$ is a very small value set to  $1e-08$.

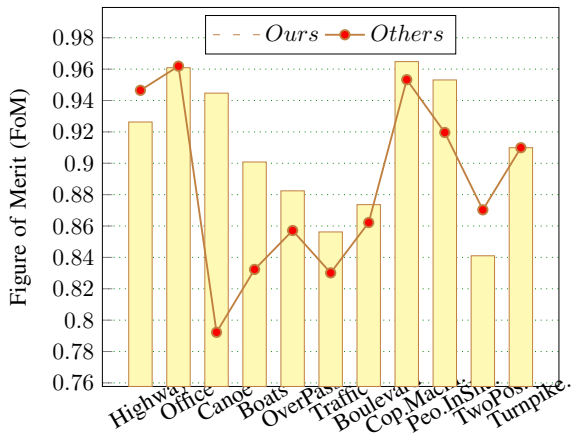
\begin{figure}[!ht]	
  \begin{center}
  	\begin{minipage}{\columnwidth}
  \resizebox{0.9\columnwidth}{!}{
	\begin{tikzpicture}
	\begin{axis}[
	symbolic x coords={Highway, Office, Canoe, Boats, OverPass, Traffic, Boulevard, Cop.Machi., Peo.InSha., TwoPos., Turnpike.},
	xtick=data,
	ylabel = {Figure of Merit (FoM)},
	ytick = {0.50,0.52,..., 1.00},
	xtick pos=left,
	ytick pos=left,
	ymajorgrids=true,
	xticklabel style = {rotate=30,anchor=north},
	legend style={at={(0.5,0.98)},
		anchor=north,legend columns=-1},
	enlarge y limits=.2]
	
	\addplot[ybar, brown,fill=yellow!80!white!45] coordinates { (Highway, 0.9264) (Office, 0.9610) (Canoe, 0.9448) (Boats, 0.9009) (OverPass, 0.8825) (Traffic, 0.8563) (Boulevard, 0.8737) (Cop.Machi., 0.9649) (Peo.InSha., 0.9532) (TwoPos., 0.8411) (Turnpike., 0.9100)};
	\addlegendentry{$Ours$}
	
	\addplot+[color=brown, thick,	mark=*,	mark options={fill=red, scale=1},text mark as node=true,	thick,	mark size=2.0pt] coordinates {(Highway, 0.9466) (Office, 0.9620) (Canoe, 0.7923) (Boats, 0.8324) (OverPass, 0.8572) (Traffic, 0.8302) (Boulevard, 0.8623) (Cop.Machi., 0.9534) (Peo.InSha., 0.9197) (TwoPos., 0.8704) (Turnpike., 0.9100)};
	\addlegendentry{$Others$}
	\end{axis}
	\end{tikzpicture}
}
\end{minipage}	
  \end{center}
  \caption{FoM vs dataset.}
  \label{fig:fom}
\end{figure}

The Table \ref{tbl:fom_vs_methods} quantitatively compares the performance of our MV-FCN with some of the results recorded in the literature from prior-art and state-of-the-art techniques. These methods include the probabilistic-based approaches as well as neural network (NN)-based learning algorithms in recent years. Figure~\ref{fig:fom}, on the other hand, summarizes the results, where the best performance of the proposed model is compared with the best results from other methods listed in Table~\ref{tbl:fom_vs_methods}. Note that, not all the models have reported results for all the datasets we tested, at the same time, there are not many literature that use NN for the task we are intended. A brief is given for the compared methods to serve as an introduction to them. The technical aspect of the methods are not analyzed, here since this work mainly revolves around the implementation of a DCNN that has the potentiality to localize the FG regions with comparable performance to the literature.


\begin{table*}[!htb]
	\begin{center}
		\begin{minipage}{\textwidth}
			\resizebox{1.0\textwidth}{!}{			
			\begin{tabular*}{40pc}{@{\extracolsep{\fill}}lcccccc@{\extracolsep{\fill}}}
			\hline
			\multirow{2}{*}{Dataset}  & \multicolumn{4}{c}{Ours} & \multicolumn{2}{c}{Others} \\
					&  S-Global &  S-Otsu &  P-Global &  P-Otsu & Prob. Models & NN Models \\
			\hline\hline
			
			{Highway} & {0.9207} & {0.8708} & \textcolor{blue}{0.9264} & {0.8790} & {0.9330\cite{Han2017}, 0.8790\cite{Varadarajan2013}, 0.9436\cite{CharlesTIP2015}} & {0.8789\cite{Zhang2015}, \textcolor{red}{0.9466}\cite{ZhaoTIP2015}} \\\hline
			
			\multirow{2}{*}{\shortstack{Office}} & \multirow{2}{*}{\shortstack{0.9127}} & \multirow{2}{*}{\shortstack{0.9574}} & \multirow{2}{*}{\shortstack{\textcolor{blue}{0.9610}}} & \multirow{2}{*}{\shortstack{0.9175}}  & \multirow{2}{*}{\shortstack{0.5864\cite{Varadarajan2013}, 0.9032 (ViBe)\cite{Xiao2016},\\ 0.9087 (RePROCS)\cite{Xiao2016}, \textcolor{red}{0.9620}\cite{CharlesTIP2015}}} & \multirow{2}{*}{\shortstack{0.9605\cite{ZhaoTIP2015}, 0.9606\cite{gemignani2015novel}}} \\
			& & & & & & \\	\hline		
								
			{Canoe} & {0.8492} & {0.8400} & {\textcolor{red}{0.9404}} & \textcolor{blue}{0.9315} & {0.7923\cite{CharlesTIP2015}, 0.6131\cite{jiang2017wesambe}} & {0.7258\cite{Zhang2015}, 0.6337\cite{ZhaoTIP2015}} \\ \hline
						
			\multirow{2}{*}{\shortstack{Boats}} & \multirow{2}{*}{\shortstack{0.8493}} & \multirow{2}{*}{\shortstack{0.8403}} & \multirow{2}{*}{\shortstack{\textcolor{red}{0.8727}}} & \multirow{2}{*}{\shortstack{\textcolor{blue}{0.8600}}} & \multirow{2}{*}{\shortstack{0.8324 \cite{Varadarajan2013}, 0.7532\cite{DBLP:BiancoCS15a}\\ 0.6932\cite{CharlesTIP2015}, 0.6401\cite{jiang2017wesambe}}} & \multirow{2}{*}{\shortstack{0.8121\cite{DBLP:BabaeeDR17},  0.6017\cite{ZhaoTIP2015},\\ 0.6560\cite{Zhang2015}}} \\
			& & & & & & \\\hline
			
			{Overpass} & \textcolor{blue}{0.8733} & {0.8128} & \textcolor{red}{0.8825} & {0.8707} & {0.6924 \cite{GuoCVPRW16}, 0.8572\cite{CharlesTIP2015}, 0.7209\cite{jiang2017wesambe}} & {0.5970\cite{ZhaoTIP2015}} \\		
			\hline	
			
			\multirow{2}{*}{\shortstack{Traffic}} & \multirow{2}{*}{\shortstack{\textcolor{blue}{0.8488}}} & \multirow{2}{*}{\shortstack{0.7683}} & \multirow{2}{*}{\shortstack{\textcolor{red}{0.8563}}} & \multirow{2}{*}{\shortstack{0.8054}} & \multirow{2}{*}{\shortstack{0.8302\cite{DBLP:BiancoCS15a}, 0.8204\cite{Han2017}, 0.7951\cite{CharlesTIP2015} \\ 0.7482\cite{Varadarajan2013}, 0.7332\cite{GuoCVPRW16}, 0.7983\cite{jiang2017wesambe}}} & \multirow{2}{*}{\shortstack{0.7750\cite{ZhaoTIP2015}, 0.8120\cite{gemignani2015novel}}} \\
			& & & & & & \\ \hline
			
			{Boulevard} & {0.7565} & {0.6816} & {\textcolor{red}{0.8737}} & {0.8116}  & {0.8174\cite{Han2017} 0.7528\cite{CharlesTIP2015}, 0.7157\cite{jiang2017wesambe}} & {\textcolor{blue}{0.8623}\cite{DBLP:BabaeeDR17}} \\\hline
		
			\multirow{2}{*}{\shortstack{CopyMachine}} & \multirow{2}{*}{\shortstack{\textcolor{blue}{0.9212}}} & \multirow{2}{*}{\shortstack{0.8998}} & \multirow{2}{*}{\shortstack{\textcolor{red}{0.9443}}} & \multirow{2}{*}{\shortstack{0.9349}} & \multirow{2}{*}{\shortstack{0.9289\cite{CharlesTIP2015}, 0.8865\cite{GuoCVPRW16}, 0.9217\cite{jiang2017wesambe} \\ 0.8171 (ViBe)\cite{Xiao2016}, 0.8535\cite{Allebosch2016}}} & \multirow{2}{*}{\shortstack{0.9534\cite{DBLP:BabaeeDR17}, 0.9039\cite{gemignani2015novel}}} \\
			& & & & & & \\ \hline
			
			{PeopleInShade} & {0.9163} & {0.8963} & {\textcolor{red}{0.9532}} & {\textcolor{blue}{0.9396}} & {0.8986 \cite{CharlesTIP2015}, 0.8976\cite{Allebosch2016}, 0.8948\cite{jiang2017wesambe}} & {0.9197\cite{DBLP:BabaeeDR17}, 0.9178\cite{gemignani2015novel}}\\
			\hline
			
			{TwoPos.PTZCam.} & {0.7953} & {0.7067} & {0.8411} & {0.8326} & {0.8285\cite{CharlesTIP2015}, 0.8315\cite{Allebosch2016}, \textcolor{blue}{0.8342}\cite{jiang2017wesambe} } & {\textcolor{red}{0.8704}\cite{DBLP:BabaeeDR17}} \\
			\hline
			
			{Turnpike\_0\_5fps} & {0.8225} & {0.8060} & {0.8946} & {\textcolor{red}{0.9100}} & {\textcolor{blue}{0.8967}\cite{CharlesTIP2015}, \textcolor{red}{0.9100}\cite{jiang2017wesambe} } & {0.4917\cite{DBLP:BabaeeDR17}} \\
			
			\hline \hline
			\end{tabular*}
		}
		\end{minipage}
	\end{center}
\caption{Performance Comparison in terms of FoM: S- training from scratch, P- pre-trained model fine-tuning, Global and Otsu stand for the two used thresholding methods. Values in \textcolor{red}{red} are the best FoM while the ones in \textcolor{blue}{blue} are the second best.}
\label{tbl:fom_vs_methods}
\end{table*}

\subsubsection{Discussion on compared methods}
In the literature \cite{Zhang2015}, Zhang~\emph{et~al.} develop a neural network (NN) that has a stacked denoising autoencoder (SDAE) learning module and a binary scene modeling based on density analysis. Whereby, the SDAE encodes the intrinsic structural information of a scene. The encoded features of image patches are then hashed in Hamming space, and then based on the hash method a binary scene is modeled through density analysis, which captures the spatiotemporal distribution information (measured by Hamming distance evaluation). Similarly, Zhao~\emph{et~al.} \cite{ZhaoTIP2015} also take advantage of NN with a stacked multilayer Self-Organizing Map (SOM) to model the BG. In which, the model is initially trained using some BG samples, and then, based on this pre-trained model, FG detection is conducted for a new test sample, at the same time the BG model is updated using the test image online as a procedure for BG maintenance. Gemignani and Rozza \cite{gemignani2015novel} extend the basic SOM model of Zhao~\emph{et~al.} with a self-balancing multi-layered SOM that tracks a long time pixel dynamics for better FG detection.
Authors in \cite{jiang2017wesambe} approach the background modeling as evidence collection of each pixel in a scene with a weight-sample-based method. They also use a minimum-weight and reward-and-penalty weighting strategy to account rapidly changing scenarios in such a way that most inefficient sample is replaced instead of the oldest sample or a random sample. Then a pixel is classified as a BG if the sum of the weights of the active samples is larger than a manually set specific threshold; otherwise, it is classified as a FG. Although their method's computational speed is quite similar to ours (2 FPS), they record a poor FG detection accuracy. On the other hand, Varadarajan~\emph{et~al.} \cite{Varadarajan2013} propose an algorithm, where the spatial relationship between pixels is taken into account by using a region-based GMM but traditional GMMs use pixel-based Gaussian distributions \cite{aki2016}.


Charles~\emph{et al.}~\cite{CharlesTIP2015} coined a system as SuBSENSE, short for Self-Balanced SENsitivity SEgmenter that adapts and integrates Local Binary Similarity Pattern (LBSP) features as additional clues to pixel intensities in a nonparametric BG model that is then automatically tuned using pixel-level feedback loops. In \cite{GuoCVPRW16}, the authors exploit Local Binary Pattern (LBP) with local Singular Value Decomposition (SVD) operator to extract invariant feature representation that is similar to the LBSP in \cite{CharlesTIP2015}. Then, they use SAmple CONsensus(SACON) approach for building the BG model based on statistics of the pixel processes (about 300 frames). Then they employ the Hamming distance, like applied in \cite{Zhang2015} with a threshold to classify each pixel as either FG or BG.These models require static and clean background samples to build up the dictionary; thus, they lack application for real-world problems. Similarly, Allebosch~\emph{et al.}~\cite{Allebosch2016} also employ local features, such as, Local Ternary Pattern (LTP) based edge descriptors and RGB color cues to classify individual pixels. They form two backgrounds based on the aforesaid edge descriptors and color cues and create two FG masks. Then, using a pixel-wise logical $AND$ operation they refine the detected FG region.
Meanwhile, Babaee~\emph{et al.} \cite{DBLP:BabaeeDR17} employ a conventional CNN, train the network with randomly selected video frames and their ground-truth segmentations patch-wise, like in \cite{Zhang2015}, and carrie out spatial-median filtering as the postprocessing of the network outputs. Although over the past two decades many algorithms have been proposed, none of them can be the ultimate model for FG inferencing. Therefore, Bianco~\emph{et al.}~\cite{DBLP:BiancoCS15a} explore a way of harnessing multiple state-of-the-art foreground detection algorithms for improving the classification accuracy. They obtain a solution tree by Genetic Programming (GP); however, this approach also cannot be acclaimed as an universal solution for the same problem.

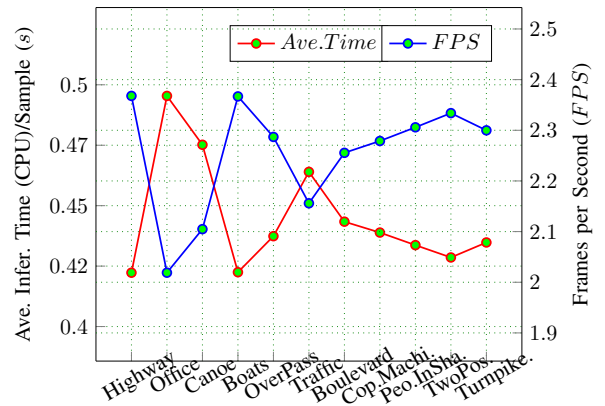
\begin{figure}[!htb]
	\begin{center}
	\begin{minipage}{\columnwidth}
		\resizebox{0.9\columnwidth}{!}{	
		\begin{tikzpicture}
		\begin{axis}[
		symbolic x coords={Highway, Office, Canoe, Boats, OverPass, Traffic, Boulevard, Cop.Machi., Peo.InSha., TwoPos., Turnpike.},
		xtick=data,
		axis y line*=left,
		ylabel = Ave. Infer. Time (CPU)/Sample ($s$),
		ytick = {0.10,0.125,..., 0.50},
		xtick pos=left,
		ytick pos=left,
		ymajorgrids=true,
		grid=major,
		xticklabel style = {rotate=30,anchor=north},
		legend style={at={(0.5,0.95)},
			anchor=north,legend columns=-1}, every node near coord,
		enlarge y limits=0.5,
		ylabel near ticks]

		\addplot[red, solid, thick, mark=*,
		mark options={fill=green, scale=1},text mark as node=true]
		coordinates { (Highway, 0.4222) (Office, 0.4953) (Canoe, 0.4751) (Boats, 0.4224) (OverPass, 0.4373) (Traffic, 0.4639) (Boulevard, 0.4433) (Cop.Machi., 0.4388) (Peo.InSha., 0.4336) (TwoPos., 0.4285) (Turnpike., 0.4347)};\label{TPF}
		\addlegendentry{$Ave. Time$}
		\end{axis}	
		
		\begin{axis}[
		symbolic x coords={Highway, Office, Canoe, Boats, OverPass, Traffic, Boulevard, Cop.Machi., Peo.InSha., TwoPos., Turnpike.},
		xtick=data,
		grid=major,
		hide x axis,
		axis y line*=right,
		ylabel = Frames per Second ($FPS$),
		ytick = {0.00,0.1,..., 3.00},
		ylabel near ticks,
		grid=major,
		legend style={at={(0.8,0.95)},
		anchor=north,legend columns=-1},
		enlarge y limits=0.5]
	
		\addplot+[solid, thick, mark=*,
		mark options={fill=green, scale=1},text mark as node=true]
		coordinates {(Highway, 2.368) (Office,2.019) (Canoe, 2.105) (Boats, 2.367) (OverPass, 2.287) (Traffic, 2.156) (Boulevard, 2.2556) (Cop.Machi., 2.279) (Peo.InSha., 2.306) (TwoPos., 2.334) (Turnpike., 2.300)};
		\addlegendentry{$FPS$}
		\end{axis}
		\end{tikzpicture}}

\end{minipage}		
	\end{center}
	\caption{Infer. Speed of the Proposed MV-FCN.}
	\label{fig:inferencing_time_per_frame}
\end{figure}

In general, most of the state-of-the-art methodologies use patch-wise processing and multi-modality based algorithms for BG establishment and a feedback-based approach as postprocessing to refine the primarily detected FG regions. Such setup ensues complex computations and higher processing time due to the time-consuming iterative pursuit of low-rank matrix or sparse matrix. On the contrary, the proposed model processes the whole input image as a single entity during inferencing. Then it refines the output by a none iterative postprocessing, resulting approximative $0.445s$ (mean average processing time) per frame, i.e., $2.25$ FPS (see Fig.~\ref{fig:inferencing_time_per_frame}) on Intel(R) Core(TM) i7-4770 CPU @ 3.40 GHz with 32.0 GB memory for FG inferencing once the network is trained. 

\section{Conclusion}\label{conclusion}
This work put an NN forwards for foreground inferencing that is inspired by recent innovations in deep learning, such as, ResNet, Inception modules, and Fully convolutional network with skip connections. The proposed model utilizes a heterogeneous set of convolutions to capture invariance features at different scales.

In traditional approaches, much time has been spent on complex mathematical modeling to optimize background generation and postprocessing to get a few more valid foreground pixels. Besides that, feature engineering and manual parameter tuning of traditional methods become unneeded since the network parameters can be learned from exemplar FG segmentations during training. For these reasons, we advocate that the proposed multi-view receptive field FCN is a novel addition to neural-based FG detection algorithms.

The qualitative and quantitative performance evaluations of the proposed MV-FCN on some challenging video sequences collected from benchmark datasets demonstrate that the model performs better than or very competitively to the prior- and state-of-the-art methods. However, the limitation of the network comes with a high number of trainable parameters. We leave this for our future direction, where we plan to optimize the network to achieve better results with less number parameters.

In application point of view, the MV-FCN can be applicable to many other applications, like MRI slice partitioning and path segmentation for autonomous vehicles. Finally, it must be considered that a perfect FG inferencing is still an intriguing task and a good FG detection system should use the knowledge derived from its ultimate purpose.


\bibliography{multiviewfcn}
\bibliographystyle{ieeetr}


\end{document}